\ifdefined\pdfminorversion\pdfminorversion=7\fi
\documentclass[conference]{IEEEtran}

\usepackage{cite}
\usepackage{amsmath,amssymb}
\usepackage{graphicx}
\usepackage{booktabs}
\usepackage{array}
\usepackage{float}

\begin{document}

\title{MDIR: A Task-Manifold Impedance Retargeting Method for Contact-Rich Teleoperation}

\author{
\IEEEauthorblockN{Liu Jiahao, Kento Kawaharazuka, Tasuku Makabe, Kei Okada}
\IEEEauthorblockA{
\textit{Department of Mechano-Informatics} \\
\textit{Graduate School of Information Science and Technology, The University of Tokyo} \\
Tokyo, Japan \\
\{liu, kawaharazuka, makabe, k-okada\}@jsk.imi.i.u-tokyo.ac.jp
}
}

\IEEEoverridecommandlockouts
\IEEEaftertitletext{%
\begin{center}
\centering
\includegraphics[width=0.95\textwidth]{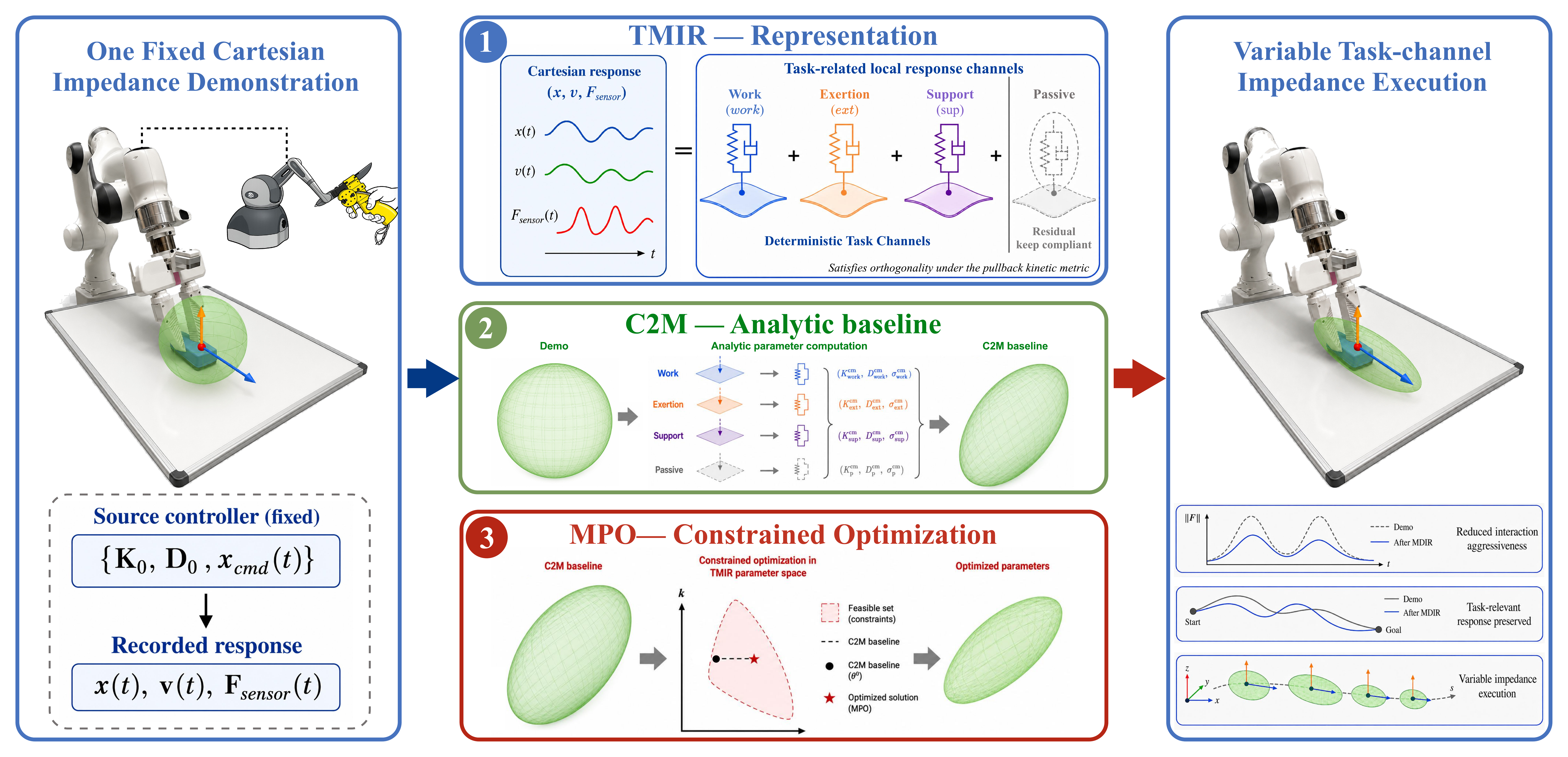}
\refstepcounter{figure}%
\label{fig:pipeline}
\parbox{0.95\textwidth}{\footnotesize Fig.~\thefigure. MDIR retargets one fixed Cartesian impedance demonstration through three functional layers. TMIR deterministically represents the recorded motion, source response, loading tendency, and contact evidence in local work, exertion, and support modes with a passive residual complement; these modes are not fixed XYZ/RPY axes. C2M analytically reconstructs an executable task-channel baseline that preserves local projected responses. MPO then refines this baseline within task-response constraints and a work-channel damping lower bound, producing a variable task-channel impedance controller with gentler contact.}
\end{center}
}

\maketitle

\begin{abstract}
Fixed Cartesian impedance makes contact-rich teleoperation demonstrations practical, but gains that secure progress and contact support also determine impact and force variability. We study single-demonstration controller-to-controller impedance retargeting. Given one fixed Cartesian impedance command sequence $\{K_0,D_0,x^{cmd}\}$, Manifold-Decomposed Impedance Retargeting (MDIR) deterministically reparameterizes the recorded controller into an executable task-channel variable-impedance command. MDIR targets this local retargeting problem by preserving projected task-channel responses near the demonstrated trajectory. It represents the source response in operational work, exertion, and support channels with a passive residual complement under a control-chain metric, computes an executable Cartesian-to-Manifold Retargeting (C2M) baseline, and applies Manifold-Constrained Parameter Optimization (MPO) to select a feasible representative with lower wrist-force peaks, impulse, force variability, and nominal controller power. Across planar wiping, pick-and-place, and pushing on a Franka Panda, the full MDIR controller passes Task Check in all 15 closed-loop executions and reduces all four aggressiveness metrics relative to the fixed-impedance demonstrations.
\end{abstract}

\section{Introduction}

Contact-rich teleoperation tasks such as wiping, pick-and-place, and pushing alternate between free motion, contact establishment, sustained loading, and release. Fixed Cartesian impedance simplifies demonstration collection, but it entangles task progress and support with impact, force accumulation, and oscillation. Reusing the demonstrated trajectory with the same controller therefore preserves both task-relevant response structure and avoidable interaction aggressiveness.

A direct remedy is to reduce every stiffness and damping gain before replay. Yet this creates a false choice: preserve the task with the original aggressive controller, or soften uniformly and risk under-driving progress or losing contact. Contact-rich reuse instead requires selective change: preserve responses that advance and support the task while relaxing low-criticality directions.

We cast this requirement as single-demonstration controller-to-controller impedance retargeting. The source is one fixed Cartesian impedance command sequence $\{K_0,D_0,x^{cmd}\}$; the target is a task-channel variable-impedance command sequence. Here, a task channel is a locally computed operational response mode, not a fixed Cartesian XYZ/RPY axis. The goal is to preserve projected task-channel responses around one demonstration while allowing geometric adaptation that reduces wrist-force maximum, impulse, force-variance upper tail, and nominal controller power.

As summarized in Fig.~\ref{fig:pipeline}, Manifold-Decomposed Impedance Retargeting (MDIR) converts a fixed Cartesian impedance demonstration through three functional layers. TMIR provides a task-structured representation comprising work, exertion (ext), and support (sup) channels with a passive residual complement. Within this representation, C2M analytically computes an executable baseline that preserves local projected responses. MPO then optimizes the TMIR parameters around this baseline under task-response constraints and a work-channel damping lower bound. The contributions are:
\begin{enumerate}
\item We formulate single-demonstration controller-to-controller impedance retargeting with local projected task-channel response as the retained object.
\item We introduce TMIR under $\Lambda_{ctrl}$ and C2M to reparameterize a fixed Cartesian impedance controller into an operational task-channel baseline.
\item We introduce MPO and validate that its task-preserving constraints reduce wrist-force aggressiveness and nominal controller power on three real contact-rich tasks.
\end{enumerate}

\section{Related Work}

Computer-controlled bilateral manipulation established an early foundation for programmable force-reflecting interaction~\cite{Inoue1971Bilateral}. Impedance, hybrid position/force, and operational-space control subsequently provided the central control ideas for contact interaction~\cite{Hogan1985Impedance,Raibert1981Hybrid,Khatib1987Operational}. They established that compliant behavior, force regulation, and task-space inertia matter as much as endpoint position. Task-priority, null-space, and Riemannian motion-policy methods later made the decomposition viewpoint explicit: task and residual directions benefit from differentiated control~\cite{Nakanishi2008Operational,Chiaverini1997SingularityRobust,Antonelli2009PrioritizedIK,Cheng2021RMPflow}. MDIR uses this geometric lesson to re-express a demonstrated fixed impedance controller under a limited task-channel equivalence relation.

Motion retargeting is the closest historical analogue because it transfers behavior across embodiment. Early methods preserve kinematic meaning or physical feasibility~\cite{Gleicher1998Retargetting,Ayusawa2017MotionRetargeting}, while teleoperation retargeting maps human commands to high-DoF embodiments~\cite{DexPilot2020,Purushottam2024ForceTeleop,Wen2025ByteDexter,Liu2025TypeTele}. These works generally preserve pose, contact geometry, or usability; MDIR instead transfers the controller representation and retains one recorded controller's local projected response.

Task-structured and compliant manipulation provide neighboring representations. Task-parameterized primitives align demonstrations with task frames~\cite{Calinon2016TaskParameterized,Huang2019KMP}, including automatically derived contact-rich frames~\cite{Mohammadi2024OptimalTaskFrame}; variable- and geometric-impedance methods learn or schedule compliance~\cite{Buchli2011LearningVIC,Kronander2014CompliantManipulation,Seo2023GIC}. MDIR instead reparameterizes one demonstrated fixed Cartesian controller with explicit response invariants: TMIR supplies the coordinates, C2M the analytic baseline, and MPO the constrained refinement.

\section{Method: Manifold-Decomposed Impedance Retargeting}

\subsection{Problem setting}

MDIR is a deterministic controller reparameterization method for one recorded fixed Cartesian impedance demonstration. We consider a demonstration collected with a fixed Cartesian impedance controller,
\begin{equation}
\mathcal{D}=\{t_k,x_k,x_k^{cmd},v_k,K_0,D_0\}_{k=1}^{N},
\label{eq:data}
\end{equation}
where $x_k \in SE(3)$ is the measured end-effector pose, $x_k^{cmd} \in SE(3)$ is the commanded pose, $v_k \in se(3)$ is the measured end-effector twist, and $K_0,D_0$ are the fixed Cartesian stiffness and damping matrices of the source controller. The corresponding Cartesian control response is
\begin{equation}
F_k^{cart}=K_0(x_k^{cmd}\ominus x_k)-D_0 v_k,
\label{eq:fcart}
\end{equation}
where $\ominus$ denotes the local pose error on $SE(3)$.

In implementation, tangent variables are local 6D end-effector twists and cotangent variables are dual 6D Cartesian wrenches. This local controller space is equipped with $\Lambda_{ctrl}$, represented by a TMIR task-channel basis, and retargeted while preserving local projected task-channel response. The resulting time-indexed controller is
\begin{equation}
\mathcal{U}^{\theta}=\{\mathcal{R}_k^{\theta}\}_{k=1}^{N},
\label{eq:targetcontroller}
\end{equation}
whose parameters vary by task channel and time:
\begin{equation}
\mathcal{R}_k=\{\mathcal{M}_k,U_k,W_k,\Theta_k\},
\label{eq:tmirtuple}
\end{equation}
where $\mathcal{M}_k$ is the local task decomposition, $U_k$ and $W_k$ are the tangent and dual-wrench bases, and $\Theta_k$ stores the channel impedances. The retargeted object is therefore an executable controller sequence, with pose evolution treated as the result of the retargeted controller.

\subsection{TMIR: Task-Manifold Impedance Representation}

TMIR first fixes the metric space in which controller coordinates are retargeted. The low-level controller computes $B=(J^T)^{\#}$ by damped SVD, with $\sigma^{\#}=\sigma/(\sigma^2+\lambda^2)$ and $\lambda=0.2$, and defines the corresponding twist-to-joint-velocity inverse as $J_{ctrl}^{\#}=B^T$. Pulling joint kinetic energy back through $\dot q=J_{ctrl}^{\#}v$ gives
\begin{equation}
\begin{aligned}
T_q&=\tfrac12\dot q^TM_q\dot q=\tfrac12v^T\Lambda_{ctrl}v,\\
\Lambda_{ctrl}&=B M_q B^T=J_{ctrl}^{\#T}M_qJ_{ctrl}^{\#}.
\end{aligned}
\label{eq:lambdactrl}
\end{equation}
Thus $\Lambda_{ctrl}$ is consistent with the implemented damped pseudoinverse and defines the controller-retargeting geometry. In contrast, $\Lambda_{phys}=(JM_q^{-1}J^T)^{-1}$ remains the physical operational-space inertia reference and metric ablation. For any positive-definite metric $G$, many bases satisfy $U^TGU=I$; TMIR selects task-semantic axes deterministically from the recorded demonstration.

The TMIR basis then defines an operational local task structure for the demonstrated controller. At each time index $k$, the task manifold is decomposed into task-response and residual subspaces,
\begin{equation}
T_{x_k}SE(3)=\mathcal{M}_{task,k}\oplus_{\Lambda}\mathcal{M}_{pass,k},
\label{eq:taskmanifold}
\end{equation}
with
\begin{equation}
\mathcal{M}_{task,k}=\mathcal{M}_{work,k}\oplus_{\Lambda}\mathcal{M}_{ext,k}\oplus_{\Lambda}\mathcal{M}_{sup,k}.
\label{eq:taskchannels}
\end{equation}
Here, $\mathcal M_{task}$ is the subspace where the source driving response exists and affects execution. The work channel is the motion-aligned positive-work channel extracted from the demonstration, ext is an active-loading wrench channel in $se^*(3)$ that is activated and validated by wrist F/T contact evidence, and support is the residual task-response channel after work and ext are removed. In contact tasks such as wiping, pushing, and placing, an environment normal provides a useful semantic reference, while ext is defined from the no-work active-loading response. By removing the work component before extracting the no-work loading response, MDIR separates progress-producing response from contact-loading response. The passive complement is the non-task residual space; its task driving response is zero by construction, although the implementation may retain weak background compliance for off-demo residual motion. Passive is treated as a residual compliance complement; the active projected-response invariant is formed by work, ext, and support.

This metric-consistent view pairs local 6D twists with dual 6D wrenches, whose natural pairing defines instantaneous power, in the implemented controller geometry.

For any constructed channel pair $(u_{i,k},w_{i,k})$, the rank-1 twist and wrench projectors are
\begin{equation}
P_{i,k}^{v}=u_{i,k}w_{i,k}^{T},\qquad
P_{i,k}^{f}=w_{i,k}u_{i,k}^{T}.
\label{eq:rankoneprojectors}
\end{equation}
The task projector sums only the active task-response channels,
\begin{equation}
\begin{aligned}
P_{task,k}^{v}&=\sum_{i\in\{work,ext,sup\}}P_{i,k}^{v},\\
P_{task,k}^{f}&=\sum_{i\in\{work,ext,sup\}}P_{i,k}^{f}
=\Lambda_{ctrl,k}P_{task,k}^{v}\Lambda_{ctrl,k}^{-1}.
\end{aligned}
\label{eq:projv}
\end{equation}

\subsection{Deterministic Task-Channel Basis Construction}

For intuition, consider planar wiping on a table. Work captures tangential task progress. Exertion captures no-work contact loading and is often approximately aligned with the table normal after the tangential work component is removed. Support retains the remaining response required to maintain pose and contact, while the passive complement provides residual compliance. The general formulation computes these local twist/wrench modes in 6D under $\Lambda_{ctrl}$.

The channel semantics---work, exertion, support, and passive residual compliance---are predefined as a common controller representation, while their instantaneous axes are computed deterministically from the recorded demonstration; users do not assign Cartesian task axes frame by frame. Work is obtained from positive-power demonstrated motion. Exertion is obtained from the no-work loading tendency and activated by contact evidence. Support is obtained from the residual source response after removing work and exertion. Work is constructed only on active-work samples where the source controller response does positive power along the demonstrated motion. Let
\begin{equation}
\nu_k=\sqrt{v_k^T\Lambda_{ctrl,k}v_k},\qquad
P^{cart}_{k}=(F_k^{cart})^Tv_k.
\label{eq:workpowergate}
\end{equation}
For $\nu_k>\epsilon_v$ and $P^{cart}_{k}>\epsilon_P$, the work direction is
\begin{equation}
\begin{aligned}
u_{work,k}&=\frac{v_k}{\nu_k},\qquad
w_{work,k}=\Lambda_{ctrl,k}u_{work,k},\\
Q_{work,k}^{cart}&=\frac{P^{cart}_{k}}{\nu_k}>0.
\end{aligned}
\label{eq:workbasis}
\end{equation}
Thus $Q_{work,k}^{cart}\dot{s}_{work,k}=P^{cart}_{k}>0$ with $\dot{s}_{work,k}=\nu_k$. Only positive-power samples enter the active-work channel; the remaining response is assigned to ext, support, or passive coordinates. Work is therefore an active geometric-evolution proxy. To separate contact loading from positive work, the ext precursor is constructed as a wrench-space no-work loading tendency,
\begin{equation}
\begin{aligned}
r_{k}^{intent}&=K_0v_k^{cmd},\\
\widetilde r_{k}^{ext}&=(I-P_{work,k}^{f})r_k^{intent},
\end{aligned}
\label{eq:extprecursor}
\end{equation}
where $v_k^{cmd}$ is obtained from consecutive commanded poses and $P_{work,k}^{f}$ removes the positive-work wrench component. After contact gating, low-pass integration, and $\Lambda_{ctrl,k}^{-1}$ normalization, this precursor gives the ext wrench axis $w_{ext,k}\in se^*(3)$ and its paired tangent direction $u_{ext,k}=\Lambda_{ctrl,k}^{-1}w_{ext,k}$. Wrist F/T sensing activates and validates sustained ext-active loading intervals, while the ext axis is constructed from the no-work loading precursor.

The support channel is extracted from the source response remaining inside $\mathcal M_{task}$ after work and ext components are removed,
\begin{equation}
F_k^{sup,raw}=(I-P_{work,k}^{f}-P_{ext,k}^{f})F_k^{cart}.
\label{eq:supprecursor}
\end{equation}
When this residual is significant, its normalized direction defines $w_{sup,k}$ and $u_{sup,k}=\Lambda_{ctrl,k}^{-1}w_{sup,k}$. This channel represents retained residual support needed for functional execution, including non-work geometric response in free motion and support response outside the explicit ext direction in contact. The ext and support bases are then orthogonalized with a $\Lambda_{ctrl,k}^{-1}$-Gram--Schmidt step, yielding separate orthogonality and normalization conditions,
\begin{equation}
w_{i,k}^{T}\Lambda_{ctrl,k}^{-1}w_{j,k}=0\ (i\neq j),
\qquad
w_{i,k}^{T}\Lambda_{ctrl,k}^{-1}w_{i,k}=1,
\label{eq:wrenchorthogonality}
\end{equation}
for $i,j\in\{work,ext,sup\}$.
The passive channel is the metric-orthogonal complement of these task channels. It provides residual background compliance for off-demo motion, while work, ext, and support form the active task-response channels.

\subsubsection{Task-channel controller synthesis}

For each rank-1 channel $i \in \{work,ext,sup\}$ in task manifold, TMIR stores a tangent basis $u_{i,k}$, the dual wrench basis $w_{i,k}=\Lambda_{ctrl,k}u_{i,k}$, and a one-dimensional impedance parameter set $(k_{i,k},d_{i,k},\delta_{i,k})$. The induced channel variables are
\begin{equation}
e_{i,k}=w_{i,k}^{T}(x_k^{cmd}\ominus x_k),\quad
\dot{s}_{i,k}=w_{i,k}^{T}v_k,
\label{eq:channelsignals}
\end{equation}
and the induced control response is
\begin{equation}
Q_{i,k}=k_{i,k}(e_{i,k}+\delta_{i,k})-d_{i,k}\dot{s}_{i,k}.
\label{eq:qi}
\end{equation}
TMIR therefore gives a task-structured representation of the demonstrated impedance behavior as an executable controller representation.

The preserved task response is reconstructed only in the task-channel dual-wrench basis,
\begin{equation}
F_k^{task}=\sum_{i\in\{work,ext,sup\}}w_{i,k}Q_{i,k},
\qquad W_{p,k}Q_{p,k}=0
\label{eq:tmirsynthesis}
\end{equation}
on the demonstrated task-response decomposition. The three rank-1 primitives define the operational response channels. A weak passive background impedance can be implemented for off-demo residual compliance; the C2M projected-response invariant is defined on the active work/ext/support channels.

The basis is complete in the metric geometry. With $P_{pass,k}^{v}=I-P_{task,k}^{v}$, any local twist can be decomposed into task and passive coordinates. On active-work samples where $v_k$ is nonzero and defines the work axis, the demonstrated velocity coordinate is assigned to the work channel by construction:
\begin{equation}
v_k=u_{work,k}\dot{s}_{work,k},\qquad
\dot{s}_{i,k}=0,\quad i\neq work.
\label{eq:twistcompleteness}
\end{equation}
Inactive, low-speed, contact-holding, and release samples are assigned to ext/support/passive coordinates or excluded from active-work RI according to their response role. This work-only velocity assignment defines the demonstrated TMIR coordinates; the original Cartesian damping can still project into ext/support because $D_0$ is generally non-diagonal in the TMIR basis. This damping leakage is the mechanical reason that C2M needs an offset compensation.

The natural tangent--cotangent pairing then reduces the demonstrated task power to the work contribution:
\begin{equation}
(F_k^{task})^Tv_k=Q_{work,k}\dot{s}_{work,k}.
\label{eq:powerdecomposition}
\end{equation}
This identity is the reason MDIR uses separate tangent and wrench projectors. It lets active work, active loading, and support be compared through their control effect under the tangent--cotangent power pairing.

\begin{figure}[!t]
\centering
\includegraphics[width=\columnwidth]{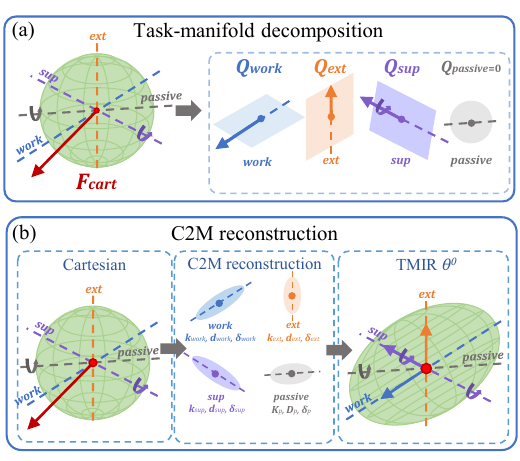}
\caption{Task-manifold decomposition and C2M reconstruction. The visualization is in task space; the representation is defined in 6D twist/wrench space.}
\label{fig:tmirc2m}
\end{figure}

\subsection{C2M: Analytic Cartesian-to-Manifold Retargeting}

C2M constructs a local projected-response TMIR baseline from the demonstrated fixed Cartesian impedance controller. It preserves projected channel responses at demonstrated states while rewriting the controller in task-manifold coordinates. The source response projected onto channel $i$ is
\begin{equation}
Q_{i,k}^{cart}=u_{i,k}^{T}F_k^{cart}
=w_{i,k}^{T}\Lambda_{ctrl,k}^{-1}F_k^{cart},
\label{eq:channelprojection}
\end{equation}
followed by
\begin{equation}
k_{i,k}^{eq}=\left(w_{i,k}^{T}K_0^{-1}w_{i,k}\right)^{-1},
\label{eq:keq}
\end{equation}
\begin{equation}
d_{i,k}^{eq}=u_{i,k}^{T}D_0u_{i,k},
\label{eq:deq}
\end{equation}
and the attractor offset that recovers the demonstrated response is
\begin{equation}
\delta_{i,k}^{c2m}=
\frac{Q_{i,k}^{cart}+d_{i,k}^{eq}\dot{s}_{i,k}}{k_{i,k}^{eq}}-e_{i,k}.
\label{eq:deltac2m}
\end{equation}
The offset in~\eqref{eq:deltac2m} is the analytic compensation that restores zero-order projected response after diagonalizing the controller in the TMIR basis. In the TMIR demonstration coordinates, all active velocity is assigned to work. The original Cartesian controller, however, uses diagonal Cartesian damping that is generally not diagonal in the TMIR basis; work velocity can therefore leak projected damping force into ext/support even when $\dot{s}_{i,k}=0$ for $i\neq work$. Thus $\delta_{i,k}^{c2m}$ absorbs this leaked projected response and recovers the demonstrated force channel while preserving the work-only demonstrated velocity assignment.
The stiffness expression is a compliance projection: it measures the displacement produced by a unit generalized force along $w_{i,k}$ and then takes its reciprocal. The damping expression preserves dissipation along the corresponding tangent direction. With the offset in~\eqref{eq:deltac2m}, the reconstructed channel satisfies
\begin{equation}
Q_{i,k}^{c2m}=Q_{i,k}^{cart},
\quad i\in\{work,ext,sup\},
\label{eq:c2mequivalence}
\end{equation}
at the demonstrated state. The baseline is therefore locally projected-response preserving and executable, and it serves as the reference point for the gentleness optimization in MPO.

Let $\mathcal A=\{work,ext,sup\}$. The baseline synthesizes executable channel-wise impedance tensors,
\begin{equation}
\begin{aligned}
K_k^{c2m}&=\sum_{i\in\mathcal A} k_{i,k}^{eq}w_{i,k}w_{i,k}^{T}
+W_{p,k}K_{p,k}W_{p,k}^{T},\\
D_k^{c2m}&=\sum_{i\in\mathcal A} d_{i,k}^{eq}w_{i,k}w_{i,k}^{T}
+W_{p,k}D_{p,k}W_{p,k}^{T}.
\end{aligned}
\label{eq:c2mrecon}
\end{equation}
Here $W_{p,k}K_{p,k}W_{p,k}^{T}$ and $W_{p,k}D_{p,k}W_{p,k}^{T}$ denote weak passive residual compliance with zero on-demo projected task response. $K_{p,k}$ and $D_{p,k}$ are set by the passive recovery-time floor; the active projected-response invariant is defined on $\mathcal A=\{work,ext,sup\}$. C2M provides zero-order projected-force equivalence at demonstrated states and preserves each active channel's local stiffness and damping response. Its equivalence is local, projected, and controller-coordinate; closed-loop execution tests how this local representative behaves under finite state deviations. Uniform Euclidean scaling instead changes all directions together and confounds progress, contact regulation, and residual compliance.

Compliance projection preserves directional elastic response and damping projection retains nonnegative dissipation along the same tangent, making C2M an interpretable executable reference for MPO.

\subsection{MPO: Manifold-Constrained Parameter Optimization}

MPO searches around the C2M baseline for a gentler representative while preserving task-channel behavior. It optimizes controller-response proxies evaluated on the recorded demonstration sequence, and closed-loop robot executions validate the resulting controller. We parameterize the target controller by
\begin{equation}
\begin{aligned}
k_{i,k}^{\theta} &= \alpha_{i,k}k_{i,k}^{c2m},\\
\delta_{i,k}^{\theta} &= \delta_{i,k}^{c2m}+\Delta \delta_{i,k},
\end{aligned}
\label{eq:theta-params}
\end{equation}
with optimization variables
\begin{equation}
\begin{aligned}
\theta_k=\{&\alpha_{work,k},\alpha_{ext,k},\alpha_{sup,k},
\zeta_{work,k},\\
&\Delta\delta_{work,k},\Delta\delta_{ext,k},\Delta\delta_{sup,k}\}.
\end{aligned}
\label{eq:theta}
\end{equation}

Because task velocity and power exchange are concentrated in the work channel, its damping is parameterized explicitly by
\begin{equation}
d_{work,k}^{\theta}=2\zeta_{work,k}\sqrt{k_{work,k}^{\theta}}.
\label{eq:workdamping}
\end{equation}
The ext and support channels use prescribed critical damping ratios, while the passive complement is set by recovery-time parameters outside the MPO active-channel variables.

MPO is a constrained representative-selection problem:
\begin{equation}
\theta^{\star}=\arg\min_{\theta\in\mathcal{H}} J_{MPO}(\theta),
\label{eq:mpo}
\end{equation}
with all objective terms normalized by the same-trial C2M or Demo scale. The objective is
\begin{equation}
\begin{aligned}
J_{MPO}&=J_{gentle}+\lambda_{contact}J_{contact}+\lambda_{work}J_{work},\\
J_{gentle}&=\sum_k\big(\|\alpha_k\|_2^2+
\lambda_{\Delta\alpha}\|\Delta\alpha_k\|_2^2\\
&\quad+\lambda_{\zeta}\zeta_{work,k}^2+
\lambda_{\Delta\delta}\|\Delta\delta_k-\Delta\delta_{k-1}\|_2^2\big),\\
J_{contact}&=\sum_{k\in\mathcal T_{stable}}\big(
\|\Delta Q_{ext,k}\|^2+\|\Delta Q_{sup,k}\|^2\big),\\
J_{work}&=\sum_{k\in\mathcal T_{work}}\big(
\|\Delta Q_{work,k}\|^2+
\lambda_{\Delta^2\delta}|\Delta^2\delta_{work,k}|^2\big).
\end{aligned}
\label{eq:mpo-objective}
\end{equation}
Here $J_{gentle}$ penalizes stiffness magnitude, gain rate, work damping, and rapid attractor changes. $J_{contact}$ suppresses jumps and upper-tail fluctuations in ext/support responses during stable contact, and $J_{work}$ penalizes fragmented progression commands.

The hard response constraints use a discrete position relative-inducer (RI). RI is a short-window induced-displacement consistency check that preserves the local task-coordinate trend induced by the reference channel response. Unless otherwise stated, $Q_i^{ref}$ is the C2M projected channel response for the same trial. Let $s_i[Q](t)$ denote the scalar displacement induced by the channel response $Q_i$ under the same control-chain unit-mass model and the recorded raw timestamps. Over a local window $W$,
\begin{equation}
\mathrm{RI}_{W}(Q_i;k)=s_i[Q](t_k)-s_i[Q](t_{k,W}^{-}),
\label{eq:ri}
\end{equation}
where $t_{k,W}^{-}$ is the earliest sample in the elapsed-time window. With $\Delta\mathrm{RI}_{i,k}=\mathrm{RI}_{W}(Q_i^\theta;k)-\mathrm{RI}_{W}(Q_i^{ref};k)$, the RI tube is
\begin{equation}
\begin{aligned}
|\Delta\mathrm{RI}_{i,k}|&\le \epsilon_{i,k}^{RI},\\
\epsilon_{i,k}^{RI}&=\rho_i
\max\!\left(|\mathrm{RI}_{W}(Q_i^{ref};k)|,\epsilon_{RI}\right),\\
Q_{sup,k}^{\theta}&=Q_{sup,k}^{ref},\quad k\in\mathcal T_{free}.
\end{aligned}
\label{eq:channelconstraints}
\end{equation}
Here $i=work$ in positive-progression work windows and $i\in\{ext,sup\}$ in stable contact; danger-transition windows are excluded. The work RI scale is formed from positive reference work response and includes a reverse-work check that preserves forward progression in active work windows.
Stable ext loading is constrained by an RMS tube. The tube preserves low-frequency active-loading magnitude only on stable-contact windows where ext is active; $R_{ext}^{ref}$ is computed from the C2M ext-channel reference over the same window. Let $R_{ext}^{\theta}(W)=\operatorname{RMS}_{W}(Q_{ext}^{\theta})$ and define $R_{ext}^{ref}(W)$ analogously:
\begin{equation}
\begin{aligned}
|R_{ext}^{\theta}(W)-R_{ext}^{ref}(W)|&\le \epsilon_{ext,k}^{RMS},\\
\epsilon_{ext,k}^{RMS}&=\rho_{ext}^{RMS}
\max\!\left(R_{ext}^{ref}(W),q_{min}^{ext}\right).
\end{aligned}
\label{eq:ext-rms}
\end{equation}
The work damping lower bound prevents the gentleness objective from reducing work-channel damping below a controllable dissipation level. Over a work-active window $W$,
\begin{equation}
\begin{aligned}
\mathcal E_D^\theta(W)&=\sum_{t_\ell\in W}
d_{work,\ell}^{\theta}\dot s_{work,\ell}^{2}\Delta t_{\ell},\\
\mathcal E_D^\theta(W)&\ge \mathcal E_{req}^{bg}(W),\qquad
\zeta_{work}^{\theta}(t_k)\ge\zeta_{work,lb}(t_k).
\end{aligned}
\label{eq:energybound}
\end{equation}
Here $\mathcal E_{req}^{bg}(W)$ is computed from the passive-safety background model as the minimum work-active dissipation required over the window, independently of the C2M work dissipation. The resulting $\zeta_{work,lb}$ is the smoothed upper envelope of an energy-budget lower bound and a recovery-time damping-ratio floor. This term serves as an engineering minimum-dissipation guard; formal passivity certification is left to future energy-safety extensions. The complete feasible set is $\mathcal{H}=\mathcal{H}_{work}^{RI}\cap\mathcal{H}_{nowork}^{RI}\cap\mathcal{H}_{sup}^{free}\cap\mathcal{H}_{ext}^{RMS}\cap\mathcal{H}_{damping}\cap\mathcal{H}_{box}$. The box set enforces $\max(k_{pass,k}/k_{i,k}^{c2m},0)\leq\alpha_{i,k}\leq1$, so optimized task-channel stiffness never falls below the passive floor.

\begin{table*}[!t]
\caption{Main closed-loop results. C2M denotes the executable analytic controller computed in the TMIR representation; MDIR denotes the final controller after MPO refinement. Task Check combines task-proxy thresholding with video safety confirmation. $^{*}$ indicates success-only proxy means for methods with failed executions. Failures are counted in Task Check and shown in Fig.~\ref{fig:taskpreservation}. $\Delta$ columns are relative to same-trial Demo; negative is lower. Best valid values are bold.}
\label{tab:main}
\centering
\scriptsize
\setlength{\tabcolsep}{3.5pt}
\renewcommand{\arraystretch}{1.05}
\resizebox{\textwidth}{!}{%
\begin{tabular}{llccccccc}
\toprule
\textbf{Task} & \textbf{Method} & \textbf{Task Check} & \textbf{Task proxy} & \textbf{Norm. pose dev. mean (\%)} & \textbf{Force max $\Delta$} & \textbf{Impulse $\Delta$} & \textbf{Force var. UT $\Delta$} & \textbf{Nominal power mean $\Delta$} \\
\midrule
Planar wiping & C2M & \textbf{5/5} & \textbf{0.906} & \textbf{0.91 $\pm$ 0.49} & +0.6 $\pm$ 18.9\% & -13.5 $\pm$ 19.4\% & -11.4 $\pm$ 33.2\% & -14.1 $\pm$ 11.2\% \\
Planar wiping & Scaling-best & 3/5 & 0.888$^{*}$ & 1.29 $\pm$ 0.53 & +6.2 $\pm$ 6.5\% & -10.9 $\pm$ 10.9\% & +3.4 $\pm$ 0.8\% & -15.7 $\pm$ 15.5\% \\
Planar wiping & MDIR & \textbf{5/5} & 0.895 & 1.40 $\pm$ 0.81 & \textbf{-8.6 $\pm$ 6.9\%} & \textbf{-29.6 $\pm$ 7.7\%} & \textbf{-47.7 $\pm$ 19.4\%} & \textbf{-24.9 $\pm$ 13.4\%} \\
\midrule
Pick-and-place & C2M & \textbf{5/5} & 2.32\,mm $\checkmark$ & \textbf{0.42 $\pm$ 0.13} & -29.5 $\pm$ 26.5\% & +5.6 $\pm$ 7.4\% & -42.5 $\pm$ 33.1\% & -2.4 $\pm$ 2.4\% \\
Pick-and-place & Scaling-best & 4/5 & 3.37\,mm $\checkmark^{*}$ & 0.54 $\pm$ 0.27 & -13.5 $\pm$ 40.8\% & +7.4 $\pm$ 18.5\% & -10.1 $\pm$ 59.8\% & +1.3 $\pm$ 2.6\% \\
Pick-and-place & MDIR & \textbf{5/5} & 2.86\,mm $\checkmark$ & 1.40 $\pm$ 0.31 & \textbf{-75.9 $\pm$ 10.5\%} & \textbf{-16.7 $\pm$ 4.6\%} & \textbf{-96.1 $\pm$ 5.0\%} & \textbf{-12.1 $\pm$ 5.4\%} \\
\midrule
Pushing & C2M & \textbf{5/5} & 2.17\,mm $\checkmark$ & \textbf{0.76 $\pm$ 0.27} & -11.3 $\pm$ 6.6\% & +6.1 $\pm$ 11.2\% & -13.1 $\pm$ 14.3\% & +2.0 $\pm$ 2.9\% \\
Pushing & Scaling-best & 3/5 & 3.81\,mm $\checkmark^{*}$ & 1.03 $\pm$ 0.45 & -7.3 $\pm$ 4.7\% & +12.1 $\pm$ 11.8\% & +7.0 $\pm$ 17.2\% & -1.0 $\pm$ 1.7\% \\
Pushing & MDIR & \textbf{5/5} & 3.60\,mm $\checkmark$ & 1.86 $\pm$ 0.56 & \textbf{-19.7 $\pm$ 7.1\%} & \textbf{-8.2 $\pm$ 2.7\%} & \textbf{-48.8 $\pm$ 24.5\%} & \textbf{-11.3 $\pm$ 4.7\%} \\
\bottomrule
\end{tabular}}
\end{table*}

Numerically, we use projected Adam with a short fixed-point projection onto the RI, RMS, damping, and box constraints, and retain the feasible candidate with the lowest $J_{MPO}$. The same optimizer settings are used across all MPO runs; the contribution is the constrained controller-retargeting formulation rather than a task-specific optimizer.

MPO can increase pose deviation because it may exploit geometric slack in the passive and non-critical directions while preserving the task-channel response set. MDIR is therefore constrained controller retargeting: it preserves the task-channel response set while allowing geometric slack in non-critical directions.

\section{Experiments}

\begin{figure}[!t]
\centering
\includegraphics[width=\columnwidth]{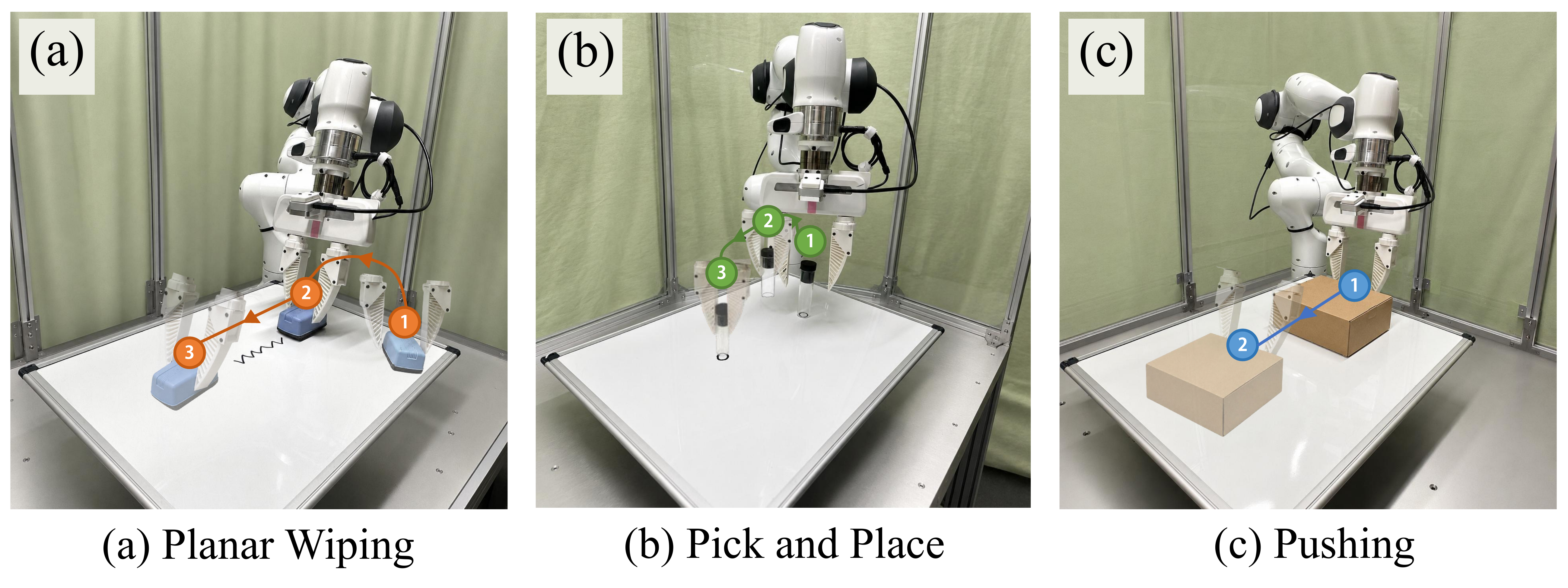}
\caption{Contact-rich manipulation tasks used in the real-robot evaluation: (a) planar wiping, (b) pick-and-place, and (c) pushing.}
\label{fig:tasks}
\end{figure}

\begin{figure}[!t]
\centering
\IfFileExists{figures/Fig_task_preservation.pdf}{%
\includegraphics[width=0.80\columnwidth]{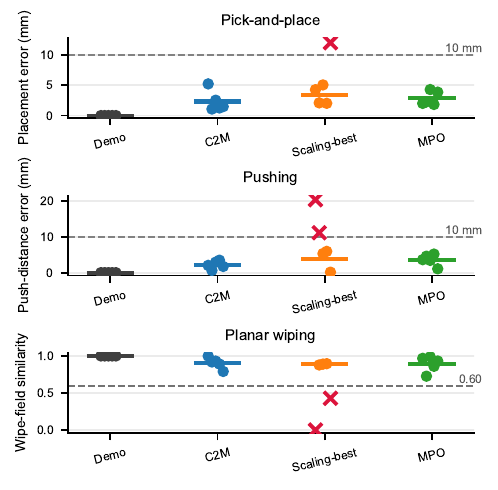}%
}{%
\fbox{\begin{minipage}[c][0.56in][c]{0.78\columnwidth}
\centering\footnotesize
Task-proxy reference placeholder.
\end{minipage}}%
}
\caption{Task-proxy reference values. Dashed lines mark proxy failure criteria; red crosses are failed executions excluded from success-only means.}
\label{fig:taskpreservation}
\end{figure}

\begin{figure*}[!t]
\centering
\includegraphics[width=0.95\textwidth]{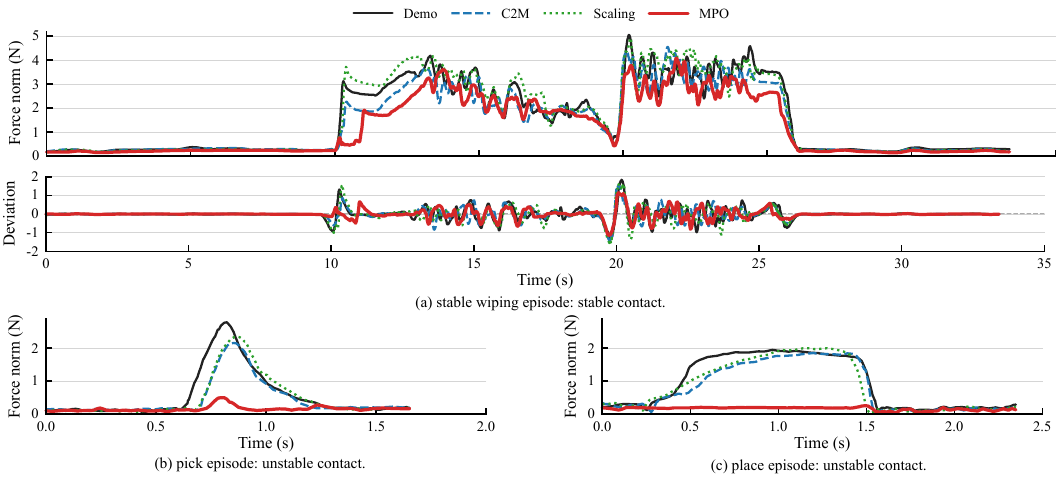}
\caption{Planar wiping comparison curves for one representative closed-loop trial ($n=1$). Pick/place panels are transition phases within this wiping trial. ``Scaling'' denotes Scaling-best. MDIR remains task-complete while reducing peaks and stable-contact variability.}
\label{fig:maincurves}
\end{figure*}

\subsection{Experimental setup}

We evaluate MDIR on a Franka Emika Panda with a Franka Hand, wrist F/T sensing, and Geomagic Touch teleoperation under a fixed Cartesian impedance controller. Each demonstration logs measured/commanded poses, executed twist, command inputs, and wrist force/torque. For each of planar wiping, pick-and-place, and pushing, we collect five independent demonstration--retargeting--execution trials. Within each trial, C2M, Scaling-best, and MDIR are generated from the same single source demonstration; demonstrations are not aggregated across trials. C2M is the executable analytic baseline represented in TMIR coordinates. Scaling-best is an oracle best-of-executed uniform-scaling baseline: for each trial, 25\%, 50\%, and 75\% candidates are executed, and the best task-preserving candidate is reported; if all fail, the trial fails.

All methods are executed on the real robot in closed loop. Task Check has two layers: quantitative task-proxy thresholds determine task outcome when available, and video safety check screens for proxy-uncovered contact loss, instability, dangerous collision, high-frequency oscillation, unsafe impact, or object-level failure. Task-specific proxies are 10\,mm placement error for pick-and-place, 10\,mm pushed-distance error for pushing, and wiping field similarity $S_{\mathrm{field}}=\sqrt{S_{\mathrm{occ}}S_{F_z}}$ with a 0.60 failure line. Figure~\ref{fig:taskpreservation} reports these proxy references. For pose deviation, $d_{SE(3)}=\sqrt{\|\Delta p\|_2^2+(0.05\theta_R)^2}$ combines translation and quaternion geodesic angle $\theta_R$; Table~\ref{tab:main} reports its time mean divided by Demo translation-path length. Force metrics use low-pass wrist-force norm, force-variance upper tail as the 95th percentile of 1.0\,s-window variance, and nominal power mean as the trajectory mean of $|F_{cmd}^{T}v|$. The 15 Demo trials span force maxima of 1.52--5.44\,N, impulses of 6.31--107\,N\,s, force-variance upper-tail values of 0.040--0.429\,N$^2$, and nominal-power means of 0.000--0.215\,W.

\subsection{Main closed-loop results}

C2M produces an executable task-channel controller in the TMIR representation and passes Task Check in all 15 executions. Starting from this analytic baseline, MPO optimizes the TMIR parameters to produce the final MDIR controller, which also passes all 15 Task Checks while reducing interaction aggressiveness. Scaling-best passes 10/15 despite its favorable oracle selection. Figure~\ref{fig:taskpreservation} lists the proxy references and red-cross failures.

MDIR reduces all four aggressiveness metrics in every task. Representative reductions are wiping impulse $29.6\pm7.7\%$, pick-and-place force maximum $75.9\pm10.5\%$, and pushing force-variance upper tail $48.8\pm24.5\%$.

Scaling-best has acceptable success-only proxy means, but its red-cross failures violate the proxy criteria. Uniform softening shows an unstable trade-off between aggressiveness reduction and task-channel preservation, even under oracle selection.

MPO optimizes constrained task-channel preservation and gentleness within MDIR, so trajectory similarity is treated as a secondary outcome. The planar-wiping $S_{\mathrm{field}}$ remains comparable to the valid baselines while all MDIR executions remain task-complete and reduce force and nominal-power aggressiveness, so the larger pose deviation is consistent with the formulation.

\subsection{Planar wiping analysis and ablation}

Planar wiping is diagnostic because it includes pickup/place transitions, stable contact, and phase changes within one execution. Figure~\ref{fig:maincurves} shows that MDIR lowers force during contact transition and stable wiping while remaining task-complete, matching the impulse and variance reductions in Table~\ref{tab:main}.

\begin{table}[!t]
\caption{Single representative planar-wiping trial used for mechanism ablation. Pose mean is in mm; force entries are max / variance-UT $\Delta$ versus Demo. Best valid values are bold.}
\label{tab:ablation}
\centering
\IfFileExists{Table_extended_ablation.tex}{%
\scriptsize
\setlength{\tabcolsep}{1.0pt}
\renewcommand{\arraystretch}{1.03}
\begin{tabular}{@{}p{0.22\columnwidth}cccccp{0.16\columnwidth}@{}}
\toprule
\textbf{Variant} & \shortstack{\textbf{Task}\\\textbf{Check}} & \shortstack{\textbf{Wipe}\\\textbf{sim.}$\uparrow$} & \shortstack{\textbf{Pose}\\\textbf{mean}$\downarrow$} & \shortstack{\textbf{Force}\\\textbf{max} $\Delta\downarrow$} & \shortstack{\textbf{Var.}\\\textbf{UT} $\Delta\downarrow$} & \shortstack{\textbf{Observed}\\\textbf{failure}} \\
\midrule
C2M & OK & \textbf{0.918} & \textbf{5.9} & -7.0\% & -27.6\% & -- \\
C2M w/o offset & Fail & 0.865 & 11.3 & +50.6\% & +181.0\% & response loss \\
Scaling-best & OK & 0.878 & 6.6 & +2.9\% & +3.2\% & -- \\
Full MDIR & OK & 0.727 & 12.5 & \textbf{-16.9\%} & \textbf{-67.5\%} & -- \\
MDIR w/o work & Fail & 0.517 & 32.2 & +110.3\% & +88.9\% & traj. drift \\
MDIR w/o ext & Fail & 0.580 & 15.7 & -16.3\% & -75.1\% & lost contact load \\
MDIR w/o support & Fail & 0.000 & 50.5 & +44.7\% & +45.9\% & lost support \\
MDIR w/o passive & Fail & 0.000 & 37.1 & -3.1\% & +8.2\% & poor residual \\
MDIR w/o $\Lambda_{ctrl}$ & Fail & 0.000 & 29.1 & +0.5\% & -11.3\% & scale/rot. loss \\
MDIR w/o MPO damp. & Fail & 0.774 & 12.2 & +13.8\% & -56.1\% & impact transient \\
MDIR w/o MPO constr. & Fail & 0.000 & 43.1 & +10.7\% & -62.5\% & divergence \\
MDIR use physmetric & Fail & 0.000 & 155.9 & +28.1\% & +144.7\% & path. startup \\
\bottomrule
\end{tabular}
}{%
\footnotesize
\setlength{\tabcolsep}{1.7pt}
\renewcommand{\arraystretch}{1.08}
\begin{tabular}{@{}p{0.28\columnwidth}ccc@{}}
\toprule
\textbf{Variant} & \shortstack{\textbf{Task}\\\textbf{Check}} & \shortstack{\textbf{Task / pose}\\\textbf{metric}} & \shortstack{\textbf{Force max /}\\\textbf{var. UT $\Delta$}} \\
\midrule
C2M & 5/5 & local baseline & valid reconstruction \\
C2M w/o offset & degraded & response loss & higher force variability \\
Full MDIR & \textbf{5/5} & 12.5 & \textbf{-19.2 / -58.3\%} \\
MDIR w/o $\Lambda_{ctrl}$ & failed & scale/rotation loss & pose/rotation error \\
MDIR channel-drop & failed & distinct drift/load loss & component-specific failure \\
MDIR w/o MPO task constr.$^{*}$ & Failed & 43.1 & -- \\
MDIR w/o MPO damp. bound$^{*}$ & Failed & 12.2 & +8.5 / +18.7\% \\
Scaling-best & 3/5 & \textbf{6.6} & -2.6 / -8.2\% \\
\bottomrule
\end{tabular}
}
\end{table}

\begin{figure}[!t]
\centering
\includegraphics[width=0.74\columnwidth]{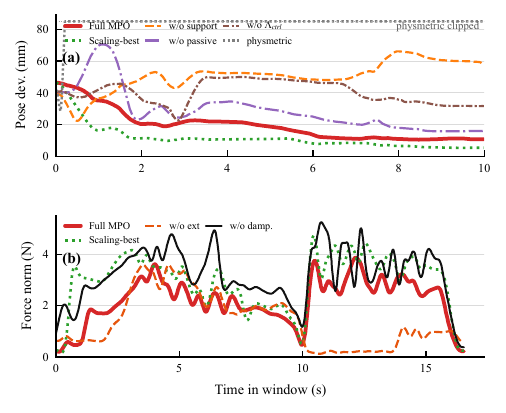}
\caption{Representative planar-wiping mechanism ablations from one closed-loop trial ($n=1$). Pose deviation exposes failures without support, passive residual compliance, or the chosen metric; wrist force exposes contact-load failures without ext or the damping bound.}
\label{fig:ablation}
\end{figure}

Table~\ref{tab:ablation} and Fig.~\ref{fig:ablation} report representative planar-wiping mechanism ablations. C2M w/o offset tests local response reconstruction. Removing work causes drift; removing ext loses wiping load; removing support or passive residual compliance increases deviation. Removing $\Lambda_{ctrl}$ causes translation--rotation scale inconsistency and loss of rotational motion retention. Using the classical physical metric makes this retargeting ill-conditioned and produced pathological startup control commands. Removing the damping bound permits impact-like transients, and removing task constraints allows divergence. Validity therefore requires the joint satisfaction of task proxies, Task Check, and sufficient damping, in addition to low force.

\section{Discussion and Conclusion}
MDIR uses TMIR as its controller representation, C2M as its analytic retargeting computation, and MPO as its constrained optimization stage. C2M yields executable analytic reconstructions; the final MDIR controller reduces every reported wrist-force and nominal-power aggressiveness metric while passing all 15 Task Checks; and the ablations show distinct failures when the offset, channels, metric, task constraints, or damping bound are removed.

MDIR's larger pose deviation is consistent with the formulation because geometric adaptation is acceptable only when constrained task-channel responses, task-level outcome metrics, and Task Check are preserved. Scaling-best exposes the converse, where uniform softening can lower force by losing progress or support. The comparison is scoped to fixed-controller retargeting; $\Lambda_{ctrl}$ is a control-chain pullback metric, and C2M equivalence is local, projected, and controller-coordinate.

Limitations remain. This study is a feasibility validation on three contact-rich tasks with five trials per task. The present formulation retargets one demonstration at a time; additional demonstrations would enlarge evaluation coverage or support future demonstration selection, but their effect on retargeting performance is not evaluated here. The current feasible set is intentionally conservative and can be narrow because task-response constraints, the damping lower bound, and parameter bounds jointly restrict adaptation. MPO therefore identifies a gentler valid representative within this set rather than exhausting the controller's full gentleness potential. The damping bound is an engineering dissipation guard; formal passivity and safety guarantees require explicit energy-safety extensions. Future work will enlarge the feasible set without sacrificing task-response preservation and extend MDIR to broader task and energy-safety conditions.

\bibliographystyle{IEEEtran}
\bibliography{refs}

\end{document}